\documentclass[10pt,twocolumn,letterpaper]{article}

\usepackage{cvpr}      

\usepackage{graphicx}
\usepackage{amsmath}
\usepackage{amssymb}
\usepackage{booktabs}
\usepackage{multirow}
\usepackage{float}

\usepackage[pagebackref,breaklinks,colorlinks]{hyperref}

\usepackage[capitalize]{cleveref}
\crefname{section}{Sec.}{Secs.}
\Crefname{section}{Section}{Sections}
\Crefname{table}{Table}{Tables}
\crefname{table}{Tab.}{Tabs.}

\def\cvprPaperID{13982} 
\def\confName{CVPR}
\def\confYear{2025}

\begin{document}

\title{Generative Modeling of Class Probability for Multi-Modal Representation Learning}

\author{JungKyoo Shin\\
Department of AI\\
Chung-Ang University\\
{\tt\small neo293@cau.ac.kr}
\and
Bumsoo Kim\\
School of CSE\\
Chung-Ang University\\
{\tt\small bumsoo@cau.ac.kr}
\and
Eunwoo Kim\thanks{Corresponding author}\\
School of CSE\\
Chung-Ang University\\
{\tt\small eunwoo@cau.ac.kr}
}

\maketitle

\begin{abstract}
Multi-modal understanding plays a crucial role in artificial intelligence by enabling models to jointly interpret inputs from different modalities. 
However, conventional approaches such as contrastive learning often struggle with modality discrepancies, leading to potential misalignments. 
In this paper, we propose a novel class anchor alignment approach that leverages class probability distributions for multi-modal representation learning. 
Our method, Class-anchor-ALigned generative Modeling (CALM), encodes class anchors as prompts to generate and align class probability distributions for each modality, enabling more effective alignment.
Furthermore, we introduce a cross-modal probabilistic variational autoencoder to model uncertainty in the alignment, enhancing the ability to capture deeper relationships between modalities and data variations.
Extensive experiments on four benchmark datasets demonstrate that our approach significantly outperforms state-of-the-art methods, especially in out-of-domain evaluations.
This highlights its superior generalization capabilities in multi-modal representation learning.
\end{abstract}

\section{Introduction}
\label{sec:introduction}

\begin{figure}[ht]
\centering
\includegraphics[width=\columnwidth]{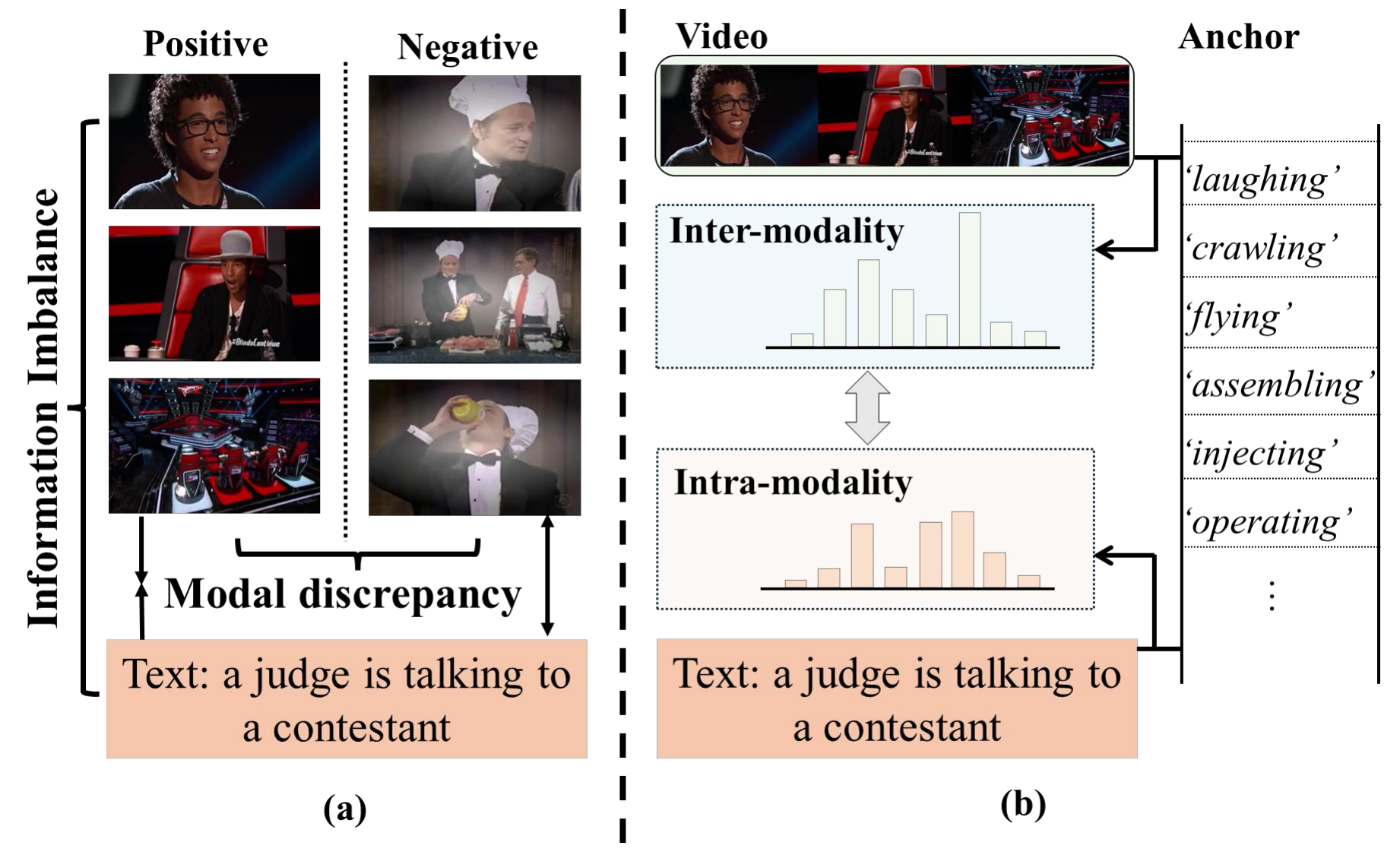}
\vspace{-1.5em}
\caption{
(a) Videos contain subtle semantic information, whereas textual descriptions often have limited expressive capacity. 
This mismatch leads to an information imbalance and modality discrepancy between video and text, resulting in the collapse of diverse video features to a limited textual representation scope. 
(b) To address this issue, we propose a class-anchor-aligned generative modeling approach. Our method generates class probability distributions by aligning prompts with inputs from each modality, effectively bridging the modality gap and preserving the diverse semantics of video content.}
\label{fig_1}
\vspace{-1em}
\end{figure}

Multi-modal understanding has emerged as a crucial area of research in deep learning, enabling models to jointly comprehend inputs from diverse modalities, such as image, audio, video, and text. 
One of the most widely used approaches in this domain is contrastive learning~\cite{radford2021learning}, which aligns features from different modalities by projecting them into a shared embedding space. 
By minimizing the distance between features of corresponding modality pairs while maximizing the distance between non-corresponding pairs in a shared latent space, this metric learning approach effectively captures semantic relationships across modalities, leading to advancements in real-world tasks, such as video-text retrieval~\cite{luo2022clip4clip} and captioning~\cite{tang2021clip4caption}.

Despite these successes, multimodal representation learning still faces significant challenges in aligning features from different modalities. 
A notable issue is the modality discrepancy and information imbalance arising from information disparity between inputs of different modalities~\cite {gabeur2020multi}. 
To address these issues, existing works for image-text alignment proposed semantic reasoning methods~\cite{li2022image,fu2023learning,gordon2024mismatchquestvisualtextual}. 
However, a video contains much more complex information, as it requires understanding both spatial information and temporal dynamics.
For instance, as illustrated in Fig. \ref{fig_1}-(a), given the text ``a judge is talking to a contestant", the modality discrepancy arises as multiple videos could match this description, each depicting different judges, contestants, and contexts.  
This leads to an information imbalance, as each video contains visual and temporal details that the brief text description cannot fully capture.

As contrastive learning approaches~\cite{luo2022clip4clip, xu2021videoclipcontrastivepretrainingzeroshot} rely on pairing each input with both its matching positive pair and non-matching negative pairs, they face significant challenges when dealing with information imbalance.
The discriminative nature of contrastive learning requires rigid definitions of positive and negative pairs. 
However, when videos and texts share only partial information, defining these pairs can overlook the inherent uncertainty and partial relationships between modalities. 
As a result, contrastive learning often fails to model the underlying data distribution effectively~\cite{liang2022gmmseg,jin2023diffusionret,wang2022understandingcontrastiverepresentationlearning}, leading to misaligned representations and hindering robust multimodal understanding.

Recent advancements in video-text retrieval have explored methods beyond contrastive learning. 
These methods aim to model the probability distribution~\cite{fang2023uatvr,jin2023diffusionret} to capture the latent characteristics of each modality better. 
Some studies adopt uncertainty-aware models to align probabilistic embeddings~\cite{fang2023uatvr} or generative models to generate joint distributions for improved modal alignment~\cite{jin2023diffusionret}. 
Although these approaches have made progress, they still depend heavily on direct matching between positive and negative pairs. 
This reliance makes them vulnerable to misalignment and limits their effectiveness in modeling the complexities of multi-modal information.

In this paper, we propose a novel Class-anchor-ALigned generative Modeling (CALM) approach to address these challenges by leveraging class anchors for enhanced cross-modal alignment.
We utilize class labels from an independent dataset that represent general categories. 
As illustrated in Fig. \ref{fig_1}-(b), each class label is transformed into a prompt, which serves as a class anchor. 
We achieve effective alignment that handles ambiguity and partial information by calculating probability distributions between each modality input and these anchors.

Our intuition is based on the observation that intra-modality relationships are less complex to model than inter-modality relationships, as latent spaces within the same modality share similar statistical features~\cite{jiang2023understandingconstructinglatentmodality, li2021universalmodelcrossmodality}. 
The probability distribution between text and the class anchor captures the abstraction of textual data. 
By aligning the video-prompt probability distribution (an inter-modal distribution) with the text-prompt probability distribution (an intra-modal distribution) over the same set of class anchors, our approach offers two advantages.
First, by introducing class anchors independent of the modality inputs, our method enriches the joint embedding space with supplementary semantic cues. 
This approach enhances the ability of the model to learn underlying data distributions and capture variations within each modality. 
Second, our approach effectively handles modality discrepancy and uncertainty. 
Our generative alignment enables the model to capture variability in inter and intra-modalities by generating a probabilistic representation of each modality. 
Unlike discriminative approaches that focus on direct pairwise comparisons~\cite{luo2022clip4clip, liu2022ts2, jin2022expectation}, the presented alignment generates intra-modal probabilistic representations from inter-modal probabilistic representations, effectively considering the distributional properties of inter and intra-modality relationships.
This approach adapts flexibly to subtle variations in video and text inputs and aligns partial or ambiguous information to enhance the generalizability of cross-modal learning.
Specifically, we augment each class label with descriptive sentences to create class anchors and generate probability distributions for each modality. 
A cross-modal probability model then aligns these distributions, forming a joint probabilistic representation between video and text.

We evaluate our model on four widely used video-text benchmarks, MSR-VTT~\cite{xu2016msr}, DiDeMo~\cite{hendricks2017localizingmomentsvideonatural}, MSVD~\cite{chen2011collecting}, and LSMDC~\cite{rohrbach2017movie}. 
We assess the performance across two tasks: video retrieval and video captioning. 
We evaluate the generalizability of our approach in both in-domain and out-of-domain scenarios. 
The results show significant improvements over the previous state-of-the-art methods, particularly in the out-of-domain scenario.
In summary, our main contributions are as follows
\begin{itemize}
    \item We propose a Class-anchor-ALigned generative Modeling (CALM) framework, introducing a novel alignment approach that leverages class anchors to bridge the modality gap.
    \item We present a cross-modal probabilistic variational autoencoder to model uncertainty in video-text alignment to capture deeper relationships between modalities.
    \item We demonstrate that the proposed method significantly outperforms existing approaches on four benchmarks, showcasing superior generalization capabilities.
\end{itemize}

\section{Related Works}
\subsection{Multi-Modal Representation Learning}
The challenge of learning joint representations from multiple modalities has been extensively studied, particularly in the context of video and text alignment. 
Our work builds upon the pretrained image-text model CLIP~\cite{radford2021learning}, taking advantage of its semantic extraction capabilities. 
Building on CLIP, existing methods focused on improving video and text representations for real-world tasks~\cite{luo2022clip4clip, liu2022ts2, gorti2022x, jin2022expectation, wang2024text, fang2023uatvr}.
CLIP4Clip~\cite{luo2022clip4clip} was the first to transfer CLIP knowledge to multi-modal tasks. 
TS2-NET~\cite{liu2022ts2} introduced a method to capture fine-grained temporal visual cues. 
X-pool~\cite{gorti2022x} used text-conditioned feature fusion across frames to enhance alignment. 
EMCL-Net~\cite{jin2022expectation} sought to bridge the modality gap using expectation maximization-based feature decomposition. 
T-MASS~\cite{wang2024text} proposed a stochastic text embedding technique to facilitate the interaction between text and video by treating text as a stochastic mass. 
UATVR~\cite{fang2023uatvr} proposed a distribution matching approach, modeling modality features as probabilistic distributions and minimizing the distance between corresponding probabilistic embeddings.
Recent approaches have focused on enhancing the pre-training process to improve the representation ability of models for video and text.
VideoCLIP~\cite{xu2021videoclipcontrastivepretrainingzeroshot} enhanced CLIP by incorporating temporal alignment for improved video-text representation learning. 
OmniVL~\cite{wang2022omnivl} introduced a unified framework capable of addressing various visual-linguistic tasks simultaneously.
Flamingo~\cite{alayrac2022flamingo} extended pre-training transformers to large visual-language models to capture cross-modal interactions with few-shot learning capabilities.
CLIP-ViP~\cite{xue2023clipvip} proposed a cross-modal learning method by generating auxiliary captions and enabling frame-wise interaction.

Despite these advancements, the aforementioned methods for pre-training and finetuning primarily focus on aligning video and text features through direct matching. 
Our approach, on the other hand, aligns probability distributions between each modality and a set of class anchors, providing a more generalizable alignment mechanism.

\subsection{Generative Models for Cross-Modal Alignment}
Generative modeling has emerged as a powerful approach for cross-modal alignment, enabling the capture of complex relationships between different modalities. 
Popular generative models, such as variational autoencoders (VAE)~\cite{kingma2013auto} and generative adversarial networks (GAN)~\cite{goodfellow2014generative} were widely used for this purpose. 
Cross-modal VAE~\cite{theodoridis2020cross} uses latent variables to learn a joint representation of text and visual data, effectively capturing the uncertainty inherent in multi-modal tasks. 
DiffusionRet~\cite{jin2023diffusionret} introduced a diffusion model to establish a shared latent space between video and text, improving alignment quality by modeling the distributions of both modalities. 
These generative modeling approaches offer flexibility in learning cross-modal interactions compared to discriminative methods~\cite{luo2022clip4clip, liu2022ts2, gorti2022x, jin2022expectation}, as they better capture variations and uncertainties in the data. 
Recent advancements have introduced even more sophisticated generative modeling approaches for cross-modal alignment~\cite{bao2023one, huang2023diffdisempoweringgenerativediffusion}. 
Uni-Diffuser~\cite{bao2023one} proposes a unified diffusion framework for multi-modal generative tasks, allowing simultaneous generation and alignment across modalities. 
DiffDis~\cite{huang2023diffdisempoweringgenerativediffusion} combines cross-modal generative and discriminative pretraining under the diffusion process. 
Machine Vision Therapy~\cite{huang2024machine} proposes a in-context learning approach that leverages large language models to identify and correct errors.

Unlike existing works generating a match probability distribution between each pair of modality inputs, the proposed method aims to align class probability distributions based on a generative model, providing a more comprehensive capture of uncertainties across modalities.

\section{Method}

\begin{figure*}[ht]
    \centering
    \includegraphics[width=0.9\textwidth]{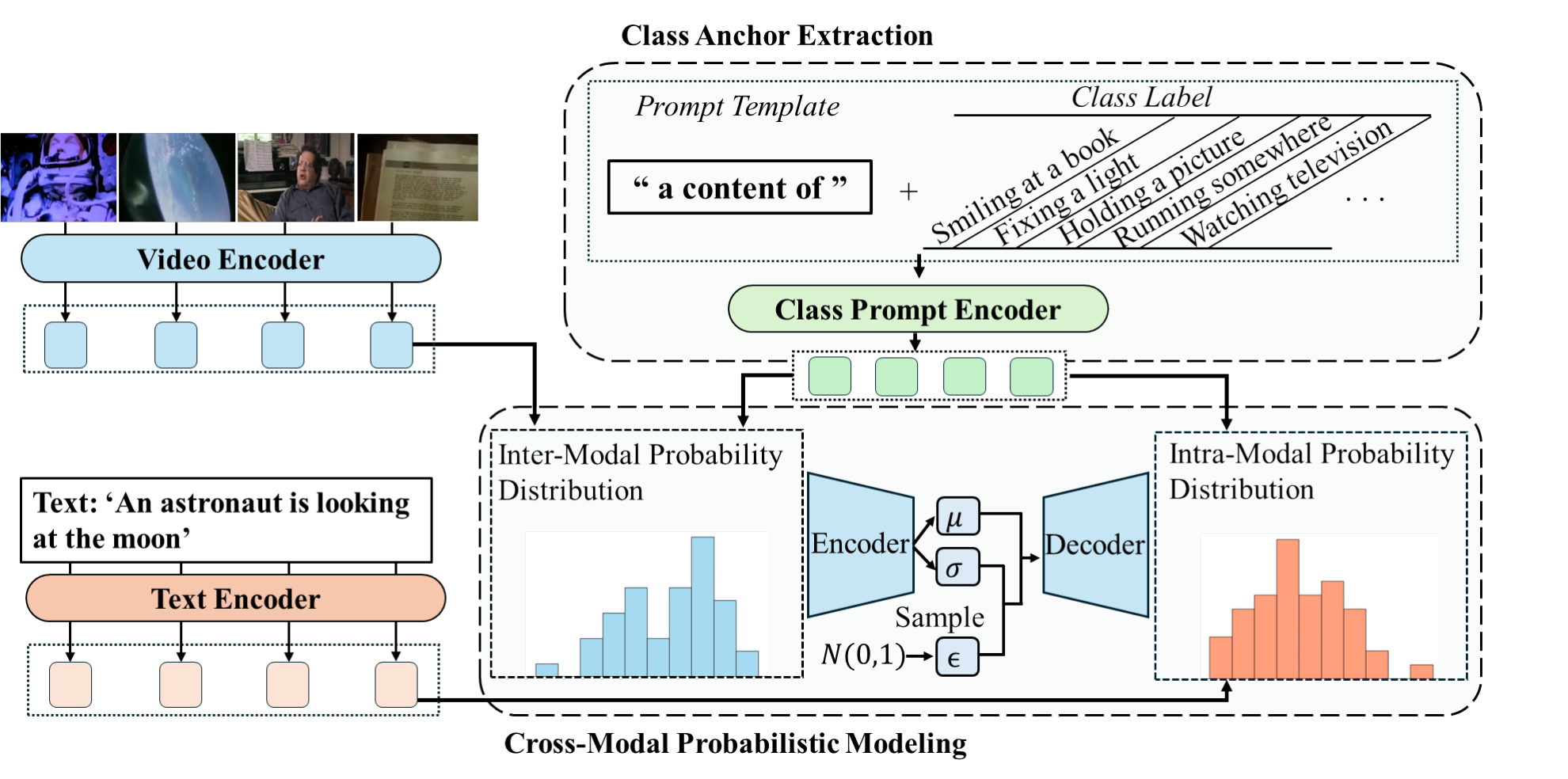}
    \caption{
    An overview of our framework. 
    We employ class labels from an independent dataset, transform them into prompts, and extract their linguistic features to serve as class anchors. 
    We then compute class probability distributions for video and text features by measuring the similarities between their features and the class anchors, effectively representing intra-modal and inter-modal relationships. 
    For modality alignment, we employ a cross-modal probabilistic variational autoencoder that takes the inter-modal probability distribution as input and reconstructs the intra-modal probability distribution to align the modalities in a shared latent space.
    }
\label{fig_2}
\vspace{-0.5em}
\end{figure*}

As illustrated in Fig.~\ref{fig_2}, our framework first leverages pre-trained CLIP encoders for feature extraction and utilizes class anchors as prompts. 
We generate probability distributions between modality inputs and prompt features and employ a cross-modal variational autoencoder for the probabilistic alignment of video and text modalities.

\subsection{Preliminary}
To extract features from videos, we uniformly sample $T$ frames as $\mathbf{v} = \{ f_1, f_2, \dots, f_T \}$. 
Each frame $f_t$ is processed by the pre-trained CLIP image encoder, $\text{CLIP}_v$.
The frame-level features are then aggregated using a temporal fusion module, $\Psi_{\text{TE}}$, resulting in a video-level feature representation as
\begin{equation}
    \mathbf{h}_t^v = \text{CLIP}_v(f_t), \quad t = 1, \dots, T , 
\end{equation}  
\begin{equation}
    \mathbf{V} = \Psi_{\text{TE}}(\mathbf{h}_1^v, \mathbf{h}_2^v, \dots, \mathbf{h}_T^v).
\end{equation}
$\mathbf{h}_t^v$ represents the visual feature of $t$-th frame and $\mathbf{V}$ represents the aggregated video feature, incorporating temporal semantics across frames.

For textual description consisting of $L$ tokens, $\mathbf{s} = \{ w_1^s, w_2^s, \dots, w_L^s \}$, we use the pre-trained CLIP text encoder, $\text{CLIP}_s$, to process each token, generating token-level features as
\begin{equation}
    \mathbf{S} = \text{CLIP}_s(\mathbf{s}) = [\mathbf{h}_1^s, \mathbf{h}_2^s, \dots, \mathbf{h}_L^s].
\end{equation}
We use the embedding of the special [CLS] token, $\mathbf{h}_{\text{CLS}}^s$, as the sentence-level representation to capture the contextual feature of the sentence.

\subsection{Class-Prompt Probability Distribution}
\paragraph{Class Anchor Extraction.} 
We define a set of $K$ class labels as $\mathbf{p} = \{ w_1^p, w_2^p, \dots, w_K^p \}$ where $w_k^p$ is the $k$-th class label obtained from an independent classification dataset $p$.
Each class label represents a general semantic category as classification datasets are typically designed to cover a wide range of categories. 
With these class labels, we adopt a prompt template~\cite{liu2021pretrainpromptpredictsystematic} for each class as
\begin{equation}
    h_k^p = \text{``The content of } [\text{label}_k] \text{''}, \quad k = 1, \dots, K.  
\end{equation}
$h_k^p$ denotes the class-specific prompt in a natural language format, and we process it through the CLIP text encoder to generate a class anchor as
\begin{equation}
    \mathbf{p}_k = \text{CLIP}_s(h_k^p) + \mathbf{e}_k^{\text{pos}},
\end{equation}
where $\mathbf{e}_k^{\text{pos}}$ is a learnable positional embedding for each prompt, allowing the model to differentiate between anchors. 
The class anchors are represented as $\mathbf{P} = [\mathbf{p}_1, \mathbf{p}_2, \dots, \mathbf{p}_K]$.

\paragraph{Probability Distribution Over Anchors.} 
To capture semantic relationships between video or text features and class anchors, we compute their cosine similarity.
We then apply a softmax function to the similarity vector to obtain the probability distribution over anchors for each modality.
The class probability distributions $\mathbf{V}_p$ for video and $\mathbf{S}_p$ for text are generated as follows:
\begin{equation}
    \mathbf{V}_p = \text{softmax}(\tau \mathbf{c}^V), \quad \mathbf{S}_p = \text{softmax}(\tau \mathbf{c}^S),
\end{equation}
where
\begin{equation}
    \mathbf{c}^V = \cos\left(\frac{1}{T} \sum_{t=1}^{T} \mathbf{h}_t^v, \mathbf{P}\right), \quad \mathbf{c}^S = \cos(\mathbf{h}_{\text{CLS}}^s, \mathbf{P}).
\end{equation}
\noindent $\tau$ is a temperature parameter that controls the sharpness of the distribution.
These distributions $\mathbf{V}_p$ and $\mathbf{S}_p$ represent the semantic relationships between the modality inputs and the class anchors. 
Specifically, $\mathbf{S}_p$ captures the intra-modal relationships between text and class anchors, while $\mathbf{V}_p$ captures the inter-modal relationships between video features and class anchors.
Since the class anchors are independent of the input data, the probabilistic distributions provide supplementary semantic cues to enrich the representations.
Aligning these distributions bridges the modality gap by mapping video and text inputs to a shared semantic space defined by class anchors. This approach captures shared and distinct semantic cues to reduce discrepancies and handle uncertainty, effectively aligning partial information across modalities.

\subsection{Cross-Modal Probabilistic Modeling}
To align the video and text modalities, we propose a cross-modal probability variational autoencoder (VAE) that reconstructs the text-anchor distribution $\mathbf{S}_p$ from the video-anchor distribution $\mathbf{V}_p$. 
It models the conditional distribution $p(\mathbf{S}_p | \mathbf{V}_p)$ via a latent variable $\mathbf{z}$ to capture semantic relationships and uncertainties between the modalities.

The encoder network encodes the video-anchor distribution $\mathbf{V}_p$ into the latent space by modeling the approximate posterior $q_\phi(\mathbf{z} | \mathbf{V}_p)$ as a Gaussian distribution with
$\boldsymbol{\mu} = f_{\text{enc}}^\mu(\mathbf{V}_p)$ and $\boldsymbol{\sigma}^2 = f_{\text{enc}}^\sigma(\mathbf{V}_p)$,
where $f_{\text{enc}}^\mu$ and $f_{\text{enc}}^\sigma$ are fully connected layers. 
Here, $\boldsymbol{\mu}$ captures the deterministic features of $\mathbf{V}_p$, and $\boldsymbol{\sigma}^2$ models the uncertainty associated with these features.
To sample $\mathbf{z}$ from the approximate posterior while allowing backpropagation, we employ the reparameterization trick~\cite{kingma2013auto}. 
We have
\begin{equation}
\mathbf{z} = \boldsymbol{\mu} + \boldsymbol{\sigma} \odot \boldsymbol{\epsilon}, \quad \boldsymbol{\epsilon} \sim \mathcal{N}(\mathbf{0}, \mathbf{I}),
\end{equation}
where $\odot$ denotes element-wise multiplication. 
The decoder network reconstructs the text-prompt distribution from the latent variable $\mathbf{z}$ as
\begin{equation}
\hat{\mathbf{S}}_p = f_{\text{dec}}(\mathbf{z}),
\end{equation}
where $\hat{\mathbf{S}}_p$ is the reconstructed probability distribution and $f_{\text{dec}}$ is a fully connected network.

To model $p_\theta(\mathbf{S}_p | \mathbf{z})$ suitable for representing probability distributions, the marginal likelihood can be expressed as
\begin{align}
\log p_\theta(\mathbf{S}_p | \mathbf{V}_p) &= \log \int p(\mathbf{S}_p, \mathbf{z} | \mathbf{V}_p) \, d\mathbf{z} \\
    &= \log \int p(\mathbf{S}_p | \mathbf{z}) p(\mathbf{z} | \mathbf{V}_p) \, d\mathbf{z}.
\end{align}
However, computing this integral directly is intractable due to the continuous latent variable $\mathbf{z}$. 
To address this, we use an approximate posterior distribution $q_\phi(\mathbf{z} | \mathbf{V}_p)$ parameterized by $\phi$.
We assume the prior $p(\mathbf{z} | \mathbf{V}_p)$ is independent of $\mathbf{V}_p$, i.e., $p(\mathbf{z} | \mathbf{V}_p) = p(\mathbf{z}) = \mathcal{N}(\mathbf{0}, \mathbf{I})$.
This assumption simplifies the computation and allows the encoder to extract modality-specific features into the latent space without being constrained by a complex prior dependent on $\mathbf{V}_p$.

Using variational inference~\cite{zhang2018advances}, we derive the Evidence Lower Bound (ELBO) for the marginal likelihood as 
$ \log p(\mathbf{S}_p | \mathbf{V}_p) \geq  \mathbb{E}_{q_\phi} \left[ \log p(\mathbf{S}_p | \mathbf{z}) \right] \nonumber \  \nonumber - \text{KL}\left( q_\phi \, \| \, p(\mathbf{z}) \right)$,
where the inequality follows the Jensen's inequality. 
The first term encourages the model to reconstruct $\mathbf{S}_p$ from $\mathbf{z}$, while the second term regularizes the approximate posterior to be close to the prior $p(\mathbf{z})$. 
While our primary focus is on modeling the conditional distribution $p(\mathbf{S}_p | \mathbf{V}_p)$, it implicitly captures the joint distribution through the shared latent variable $\mathbf{z}$. 
By modeling the relationship between each modality and $\mathbf{z}$, the model learns a joint representation that expresses the shared semantics between video and text, enhancing the alignment and capturing deeper relationships.

\subsection{Training Objective}
Maximizing the ELBO yields the training objective, comprising two main components: the reconstruction loss and the KL divergence regularization term.
The reconstruction loss, $\mathcal{L}_{\text{rec}}$, guides the model to generate probability distribution $\hat{\mathbf{S}}_p$ closer to $\mathbf{S}_p$, and is expressed as
\begin{equation}
\mathcal{L}_{\text{rec}} = - \mathbb{E}_{q_\phi} \left[ \log p_\theta(\mathbf{S}_p | \mathbf{z}) \right] \approx - \sum_{k=1}^{K} \mathbf{S}_p^{(k)} \log \hat{\mathbf{S}}_p^{(k)},
\end{equation}
where $\hat{\mathbf{S}}_p^{(k)}$ is the predicted probability for the class anchor $k$, and $\mathbf{S}_p^{(k)}$ is the true probability from the text modality.
The KL divergence regularizes the approximate posterior to be close to the prior, preventing overfitting and encouraging a smooth latent space as
\begin{align}
\mathcal{L}_{\text{KL}} &= \text{KL}\left( q_\phi \, \| \, p(\mathbf{z}) \right)       \\ 
&= \int q_\phi(\mathbf{z} \mid \mathbf{V}_p) \left[ \log q_\phi(\mathbf{z} \mid \mathbf{V}_p) - \log p(\mathbf{z}) \right] d\mathbf{z}  \\ 
&\approx \frac{1}{2} \sum_{i=1}^{d} \left( \mu_i^2 + \sigma_i^2 - \log \sigma_i^2 - 1 \right),
\end{align}
where $d$ is the dimensionality of the latent space.

Our final loss for multi-modal representation learning is as follows
\begin{equation}
\label{eq:total_loss}
\mathcal{L} = \mathcal{L}_{\text{rec}} + \alpha \mathcal{L}_{\text{KL}} + \mathcal{L}_{\text{task}},
\end{equation}
where $\alpha$ is the hyperparameter balancing the reconstruction loss and the KL divergence. 
$\mathcal{L}_{\text{task}}$ represents an objective function for a multi-modal task.

\section{Experiments}
\subsection{Datasets}
We evaluate CALM using four benchmark datasets.
\textbf{MSR-VTT}~\cite{xu2016msr} contains 10,000 online video clips (10-30 seconds each), annotated with 200,000 captions across various topics.
\textbf{LSMDC}~\cite{rohrbach2017movie} includes 118,081 clips from 200 movies, each paired with a sentence from scripts and audio descriptions.
\textbf{DiDeMo}~\cite{hendricks2017localizingmomentsvideonatural} comprises 10,642 personal videos with 40,543 temporally localized sentences.
\textbf{MSVD}~\cite{chen2011collecting} features 1,970 online video clips, each with about 40 diverse English annotations.

\begin{table*}[t]
\centering
\caption{Experimental results of video retrieval trained on MSR-VTT. ``$\rightarrow$'' indicates the out-of-distribution evaluation. }
\vspace{-0.5em}
\label{table_3}
\begin{tabular}{l|llll|llll|llll} \hline
\multirow{2}{*}{}              & \multicolumn{4}{c|}{ MSR-VTT} & \multicolumn{4}{c|}{$\rightarrow$ DiDeMo} & \multicolumn{4}{c}{$\rightarrow$ LSMDC} \\ \cline{2-13} 
             & R@1   & R@5   & R@10  & MnR & R@1   & R@5  & R@10  & MnR  & R@1   & R@5  & R@10 & MnR \\ \hline
CLIP4Clip~\cite{luo2022clip4clip}    & 43.1  & 72.7  &  81.5 & 15.7& 32.4  & 59.9 & 68.8  & 37.9 & 15.9  & 30.5 & 38.3 & 115.1\\
EMCL~\cite{jin2022expectation}         & 47.8  & 73.5  &  83.6 & 13.6& 34.2  & 59.7 & 69.5  & 31.8 & 15.2  & 27.7 & 36.5 & 119.7\\
UATVR~\cite{fang2023uatvr}        & 47.0  & 73.1  &  83.5 & 12.7& 24.9  & 44.9 & 53.8  & 79.1 & 10.7  & 21.9 & 28.6 & 153.1\\ 
DiffusionRet~\cite{jin2023diffusionret} & 49.0  & 75.2  &  82.7 & 12.1& 36.8  & 65.0 & 72.1  & 30.9 & 17.2  & 33.0 & 39.7 & 110.3  \\
T-MASS~\cite{wang2024text}       & 48.9  & 76.3  &  85.3 & \textbf{11.7}& 37.3  & 64.8 & 74.2  & 26.3 & 19.6  & 37.0 & 46.4  & 87.5  \\ \hline
CALM (Ours)         & \textbf{50.8}  & \textbf{77.5}  &  \textbf{85.8} & \textbf{11.7}& \textbf{41.2}  & \textbf{66.3} & \textbf{76.3}  & \textbf{16.1} & \textbf{21.4}  & \textbf{39.7} & \textbf{47.8} & \textbf{80.9}  \\ \hline
\end{tabular}
\vspace{-0.5em}
\end{table*}

\begin{table*}[t]
\centering
\caption{Experimental results of video retrieval trained on DiDeMo.}
\vspace{-0.5em}
\label{table_4}
\begin{tabular}{l|llll|llll|llll} \hline
\multirow{2}{*}{}              & \multicolumn{4}{c|}{DiDeMo} & \multicolumn{4}{c|}{$\rightarrow$ LSMDC} & \multicolumn{4}{c}{$\rightarrow$ MSR-VTT} \\ \cline{2-13} 
             & R@1   & R@5   & R@10  & MnR & R@1   & R@5  & R@10  & MnR  & R@1   & R@5  & R@10 & MnR \\ \hline
CLIP4Clip~\cite{luo2022clip4clip}    & 42.8 & 68.5 & 79.2  & 18.9  & 14.5  & 28.8 & 36.2  & 119.4& 35.4& 59.0 & 69.4   & 25.5 \\
EMCL~\cite{jin2022expectation}         & 47.8  & 74.1  & 84.8  & 12.2   & 14.3   & 27.7  &  35.1& 127.8  & 34.3   & 60.6   &  71.4    & 26.7 \\
UATVR~\cite{fang2023uatvr}        & 43.1  & 71.8  & 82.3 & 15.1   &  9.9 & 20.5 &  26.9 & 171.6  &  30.2  & 53.0  & 63.1 & 35.8    \\ 
DiffusionRet~\cite{jin2023diffusionret} & 46.6  & 74.7  & 82.7 & 14.3   &  13.9 & 26.1 & 34.2 & 125.6  & 31.7  & 56.4 & 67.2 & 30.6  \\ 
T-MASS~\cite{wang2024text}       & 46.7  & 73.1  &  82.4 & 14.2   & 20.4  & 37.8  & 46.0  & 81.3 & 39.7 & 66.3 & 76.6 & 17.4  \\ \hline
CALM (Ours)         & \textbf{51.1}  & \textbf{77.3}  & \textbf{84.2}  & \textbf{12.8}   & \textbf{22.1} & \textbf{40.5} & \textbf{48.7}   & \textbf{78.9}  & \textbf{41.7}  & \textbf{66.5} & \textbf{79.0}  & \textbf{16.0}  \\ \hline
\end{tabular}
\vspace{-0.5em}
\end{table*}

\begin{table*}[ht!]
\centering
\caption{Experimental results of video retrieval trained on LSMDC.}
\vspace{-0.5em}
\label{table_2}
\begin{tabular}{l|llll|llll|llll} \hline
\multirow{2}{*}{}              & \multicolumn{4}{c|}{ LSMDC} & \multicolumn{4}{c|}{$\rightarrow$ DiDeMo} & \multicolumn{4}{c}{$\rightarrow$ MSR-VTT} \\ \cline{2-13} 
             & R@1   & R@5   & R@10 & MnR & R@1   & R@5  & R@10  & MnR  & R@1   & R@5   & R@10 & MnR \\ \hline
CLIP4Clip~\cite{luo2022clip4clip}    & 22.6  & 41.0  & 49.1 & 61.0& 30.1  & 55.2 & 66.0  & 37.7 & 28.9  & 53.6  & 62.7 & 33.6\\
EMCL~\cite{jin2022expectation}         & 22.4  & 40.6  & 49.2 & 58.8 & 29.0   & 55.4  & 64.8 & 35.8  & 30.1   & 53.4   & 63.4     & 39.9 \\
UATVR~\cite{fang2023uatvr}        & 21.9  & 41.4  & 49.3 & 58.3 & 30.5   & 55.9  & 66.2 & 36.1  & 30.4   & 53.8   & 64.2  & 34.9 \\ 
DiffusionRet~\cite{jin2023diffusionret} & 24.4  & 43.1  & 54.3 & \textbf{40.7} & 31.5   & 57.9  & 67.1 & 32.5  & 30.4   & 53.5   & 65.6   & 33.4 \\ 
T-MASS~\cite{wang2024text}       & 26.0  & 47.5  & \textbf{56.4} & 46.2 & 27.2  & 51.9 & 63.1  & 34.9 & 28.5  & 52.9 & 65.2 & 31.3  \\ \hline
CALM (Ours)         & \textbf{27.5}  & \textbf{47.9}  & 56.3 & 45.4 & \textbf{33.4}   & \textbf{59.3}  & \textbf{68.3} &\textbf{31.9}  & \textbf{32.5}  & \textbf{55.5}  & \textbf{66.2}   & \textbf{30.2}   \\ \hline
\end{tabular}
\vspace{-1em}
\end{table*}

\subsection{Tasks and Metrics}
We assess CALM on two tasks, video retrieval and video captioning.
\textbf{Video Retrieval} retrieves relevant videos based on textual queries, measured by Recall at N (R@N) indicating retrieval accuracy within top N results, and mean rank (MnR) for overall retrieval performance.
\textbf{Video Captioning} generates descriptive sentences for videos, evaluated using BLEU-4~\cite{papineni2002bleu} (B),  ROUGE-L~\cite{lin-2004-rouge} (R), METEOR~\cite{banerjee2005meteor} (M) and CIDEr~\cite{vedantam2015cider} (C) metrics, assessing n-gram overlap, semantic relevance, and consensus with human annotations, respectively. 

\subsection{Implementation Details}
We use class labels from the Charades dataset~\cite{sigurdsson2016hollywoodhomescrowdsourcingdata} as class anchors, providing diverse, dynamic action and event categories.
The number of class labels, $K$, is set to 157, ensuring broad category coverage.
We implement our CALM framework using the ViT-B/32 CLIP model~\cite{radford2021learning}. 
In the cross-modal probabilistic modeling, we set the dimensionality of the latent space $d$ to 256. 
The hyperparameter $\alpha$ in the loss function is set to 0.1. 
We train our models using the AdamW optimizer~\cite{loshchilov2017decoupled} with a learning rate of $10^{-5}$. 
We employ a batch size of 128 and train for five epochs for the retrieval task and 20 epochs for the captioning task. 
To prevent overfitting, we apply dropout with a rate of 0.1 in the encoder and decoder networks of the VAE.
For video processing, we uniformly sample 12 frames from each video and resize them to $224 \times 224$.
For a fair evaluation, we re-implement existing methods that build upon a pretrained model (CLIP)~\cite{luo2022clip4clip, jin2022expectation, fang2023uatvr, jin2023diffusionret, wang2024text, perez2021improving, gu2023textknowledgegraphaugmented, shen2024accuratefastcompressedvideo, tang2021clip4caption} and evaluate their performance under our experimental setup. 
Note that results from our re-implemented models may differ from the original papers due to variations in training and evaluation settings. 
While some methods evaluate only at the end of each epoch, others evaluate multiple times within each epoch.
To ensure a fair comparison, we standardize the procedure by evaluating all methods at the end of each epoch.

\subsection{Performance Comparison}
We evaluate the effectiveness of CALM by comparing it with state-of-the-art methods for video retrieval on three datasets (MSR-VTT, DiDeMo, and LSMDC) and video captioning on two datasets (MSR-VTT and MSVD).
We evaluate on in-domain and out-of-domain scenarios to assess the generalization capabilities of our model. 
In the out-of-domain evaluation, we train the model on the training set of one dataset and measure its performance on the test set of another dataset, which remains unseen during training.

\subsubsection{Video Retrieval Results}

\noindent  \textbf{In-Domain Evaluation.} 
CALM consistently outperforms existing state-of-the-art methods across all datasets and metrics, demonstrating its effectiveness in capturing cross-modal relationships. 
Specifically, on the MSR-VTT dataset (Table~\ref{table_3}), CALM attains an R@1 of 50.8\%, surpassing the previous best result of 49.0\% from DiffusionRet by a margin of 1.8\%. 
On DiDeMo (Table~\ref{table_4}), CALM achieves an R@1 of 51.1\%, exceeding the previous best result of 47.8\% by EMCL by 3.3\%. 
For LSMDC (Table~\ref{table_2}), CALM achieves an R@1 of 27.5\%, improving the result of T-MASS by 1.5\%.

\noindent \textbf{Out-of-Domain Evaluation.} 
To assess generalization performance, we evaluate the models in out-of-domain scenarios and report the results in Tables~\ref{table_3}, \ref{table_4} and \ref{table_2}.
We study the performance results from two perspectives: the performance increase of our method compared to previous works and the average performance drop from in-domain to out-of-domain evaluations.
When trained on MSR-VTT, DiDeMo, and LSMDC, CALM achieves average performance improvements of 2.9\%, 1.9\%, and 2.0\% in R@1, respectively. 
For the average performance drop, we calculate it as the performance decrease in R@1 from in-domain to out-of-domain evaluations.
CALM exhibits drops of 13.7\%, 13.8\%, and 5.8\% when evaluated on the test sets of MSR-VTT, DiDeMo, and LSMDC, respectively. 
For MSR-VTT and DiDeMo, CLIP4Clip shows slightly lower average performance drops of 11.0\% and 11.6\%, respectively. 
However, this may be attributed to the comparatively lower in-domain performance of CLIP4Clip. 
For LSMDC, CALM shows the lowest average performance drop.
Overall, CALM demonstrates superior generalization abilities, with higher R@1 scores and competitive average performance drops compared to other methods. 
This suggests that our model learns more generalized representations due to the introduction of class anchors and probabilistic modeling.

\begin{table}[t]
\centering
\caption{Experimental results of video captioning trained on MSR-VTT.
``MSVD $\rightarrow$ MSR-VTT" indicates out-of-domain results, where the model is trained on MSVD and evaluated on MSR-VTT.}
\vspace{-0.5em}
\label{table_5}
\begin{tabular}{lllll} \hline
MSR-VTT                   & B           & M             & C     &  R         \\  \hline
SemSynAN~\cite{perez2021improving}                   & 46.4          & 30.4          & 51.9  &  54.7          \\
TextKG~\cite{gu2023textknowledgegraphaugmented}            & 43.7          & 29.6          & 52.4  &  62.4           \\
CoCAP~\cite{shen2024accuratefastcompressedvideo}                    & 43.1          & 29.8          & 56.2  & 62.7          \\
CLIP4Caption~\cite{tang2021clip4caption}                  & 46.1          & 30.7          & 57.0  &  64.1          \\
CALM (Ours)                      & \textbf{47.8} & \textbf{31.1} & \textbf{59.3}  &  \textbf{65.0} \\ \hline
MSVD $\rightarrow$ MSR-VTT & B           & M             & C     &  R         \\ \hline
CoCAP~\cite{shen2024accuratefastcompressedvideo}                    & 28.1          & 22.6          & 29.1  & 54.8            \\
CLIP4Caption~\cite{tang2021clip4caption} & 30.1         & 23.5          & 30.5  & 55.6           \\ 
CALM (Ours)                      & \textbf{34.2}          & \textbf{26.1}  & \textbf{35.6} &  \textbf{58.9} \\ \hline
\end{tabular}
\vspace{-0.5em}
\end{table}

\noindent \textbf{Complexity and Time.}
Our approach slightly increases computational cost compared to the baseline, T-MASS, adding only 0.5M parameters (from 152.6M to 153.1M) and 0.08 seconds to the training time per batch (from 1.06s to 1.14s). This minimal overhead results from the shallow encoder-decoder structure of our VAE. Importantly, the cross-modal probabilistic modeling used during training enhances modal alignment without introducing extra computational costs during inference.

\subsubsection{Video Captioning Results}
As shown in Tables~\ref{table_5} and~\ref{table_6}, our model achieves significant improvements over state-of-the-art methods on both the MSR-VTT and MSVD datasets. 
For in-domain evaluations on MSR-VTT, CALM surpasses existing methods by 1.4, 0.4, 2.3, and 0.9 in BLEU-4, METEOR, CIDEr, and ROUGE-L metrics, respectively. On the MSVD dataset, CALM demonstrates comparable performance to SemSynAN~\cite{perez2021improving} in BLEU-4, METEOR, and ROUGE-L, while achieving a substantial increase of 2.7 in CIDEr, indicating better alignment with human annotations.
In our out-of-domain evaluations, we compare CALM with CLIP4Cap and CoCAP as these methods have achieved the highest CIDEr scores in in-domain settings among existing approaches, providing strong baselines for assessing the performance of our method.
On MSR-VTT trained on MSVD, CALM outperforms these methods with improvements of 4.1, 2.6, 5.5, and 3.3 in BLEU-4, METEOR, CIDEr, and ROUGE-L, respectively. 
Similarly, for MSVD trained on MSR-VTT, CALM attains gains of 4.3, 1.3, 0.5, and 1.2 in the same metrics. 
These consistent gains across different metrics, datasets, and evaluation scenarios underscore the ability to handle modality discrepancies and information imbalance effectively, capturing nuanced semantic relationships between video and text and generating high-quality captions with strong generalization performance.

\begin{table}[t]
\centering
\caption{Experimental results of video captioning trained on MSVD. ``MSR-VTT $\rightarrow$ MSVD" indicates out-of-domain results.}
\vspace{-0.5em}
\label{table_6}
\begin{tabular}{lllll} \hline
MSVD                     & B           & M             & C              &  R         \\ \hline
SemSynAN~\cite{perez2021improving}                 & 64.4          & \textbf{41.9}          & 111.5          &  \textbf{79.5}          \\
TextKG~\cite{gu2023textknowledgegraphaugmented}    & 60.8          & 38.5          & 105.2          & 75.1              \\
CoCAP~\cite{shen2024accuratefastcompressedvideo}                  & 55.9        & 39.9   & 113.0         & 76.8  \\
CLIP4Caption~\cite{tang2021clip4caption}            & 63.0            & 40.6          & 114.6          &77.3          \\
CALM (Ours)                     & \textbf{64.4} & 41.2 & \textbf{115.7} & 77.8 \\ \hline
MSR-VTT $\rightarrow$ MSVD & B           & M             & C              &  R         \\ \hline
CoCAP~\cite{shen2024accuratefastcompressedvideo}                  & 39.1        & 34.3      & 69.8         &  69.8           \\
CLIP4Caption~\cite{tang2021clip4caption}               & 39.9        & 32.9         & 67.8          &    63.7     \\
CALM (Ours)                     & \textbf{44.2}          & \textbf{35.6 }        & \textbf{70.3 }         &  \textbf{71.0 } \\ \hline
\end{tabular}
\vspace{-1em}
\end{table}

\begin{figure*}[ht]
\centering
\includegraphics[width=0.96\textwidth]{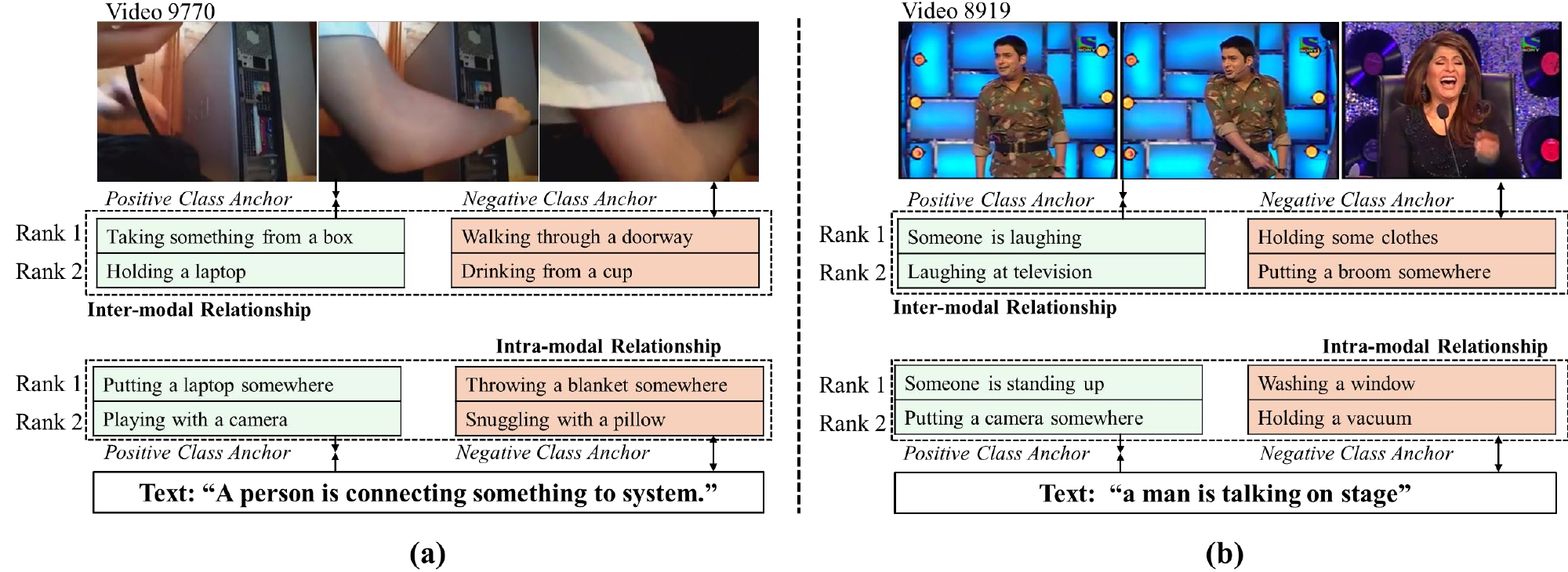}
\vspace{-1em}
\caption{Qualitative video retrieval results on the MSR-VTT dataset. 
Selected anchors capture distinct semantic cues, either aligning shared content (a) or highlighting complementary information to address modality imbalance (b).
Inter-modal and intra-modal relationships serve as supplementary semantic cues, enhancing the semantic alignment between video and text and improving retrieval performance.}
\vspace{-1em}
\label{fig_3}
\end{figure*}

\begin{table}[t]
\centering
\caption{Comparison of the number of class anchors on video retrieval performance on MSR-VTT.}
\vspace{-0.5em}
\label{table_class_prompts}
\begin{tabular}{l|l|ccc} \hline
Class anchors & Type & R@1 & R@5  & MnR \\ \hline
0 (Baseline) & Charades & 48.9 & 76.3  & 11.7 \\
50  & Charades   & 48.1 & 74.0 &  12.1 \\
100  & Charades & 49.7 & 76.5 & \textbf{11.5} \\ 
157   & Charades & 50.8 & 77.5 &  11.7 \\ \hline
91    & COCO & 50.3 & 76.0 & 11.7 \\ 
258  & Charades+COCO & 51.1 & 75.7 & 10.7 \\ \hline
\end{tabular}
\vspace{-1em}
\end{table}
\subsection{Ablation Analysis}
\noindent \textbf{Effect of Class Anchors.} 
Table~\ref{table_class_prompts} presents the impact of varying the number and quality of class anchors on video retrieval performance using the MSR-VTT dataset.
We randomly select class anchors, train and evaluate performance with 0 anchors as the baseline, followed by 50, 100, and 157 anchors. 
When using 50 anchors, the model achieves an R@1 of 48.1\%, slightly below the baseline (48.9\%), suggesting that a limited anchor set can introduce bias and negatively impact performance. 
Increasing to 100 anchors raises performance to an R@1 of 49.7\%, indicating that greater diversity among anchors improves semantic alignment. 
Employing the complete set of 157 anchors yields an even higher R@1 of 50.8\%, demonstrating the advantage of comprehensive anchor coverage for capturing richer semantic information.
Additionally, we replace Charades anchors with 91 object-centric COCO image labels~\cite{lin2014microsoft}, which lack explicit descriptions of actions or events. 
Despite this domain mismatch, employing COCO anchors still outperform the baseline. This suggests that the model effectively adapts to anchor relevance, using even seemingly irrelevant anchors to guide modal alignment. 
Further, combining Charades and COCO anchors lead to the best performance improvements in R@1, emphasizing that the model benefits significantly from integrating both domain-specific and general semantic cues.

\noindent \textbf{Generative Modeling.} 
In Table~\ref{table_generative_learning}, we study the effect of our generative modeling by replacing the variational autoencoder with discriminative loss functions, such as KL divergence, cross-entropy, and mean squared error (MSE) losses between the class probability distributions. 
By minimizing the distance between the class probability distributions of the video and text modalities, these loss functions align them without employing our approach.
CALM, employing the variational autoencoder, achieves the highest R@1 of 50.8\%, outperforming the discriminative methods. 
It also outperforms those methods in out-of-domain evaluation, achieving the highest R@1 of 41.2\%. 
This underscores the effectiveness of our VAE-based method in modeling uncertainty and capturing the underlying data distribution, resulting in improved cross-modal alignment.


\begin{table}[t]
\centering
\caption{Comparison of generative and discriminative learning approaches on video retrieval performance on MSR-VTT for in-domain and DiDeMo for out-of-domain evaluation.}
\vspace{-0.5em}
\label{table_generative_learning}
\begin{tabular}{l|cc|cc} \hline
 & \multicolumn{2}{c|}{ MSR-VTT}  &  \multicolumn{2}{c}{$\rightarrow$ DiDeMo} \\ \hline 
Loss &  R@1 &  R@5 & R@1 & R@5  \\ \hline 
Baseline                  & 48.9 & 76.3 & 37.3 & 64.8   \\
KL Divergence Loss        & 49.5 & 75.0 &  38.8 & 65.7 \\
Cross-Entropy Loss        & 50.1 & 75.5 & 38.3 & 65.5\\
MSE Loss                  & 48.7 & 74.2 & 37.3  & 65.4 \\ \hline
CALM (Ours)          & \textbf{50.8} & \textbf{77.5} & \textbf{41.2} & \textbf{66.3}   \\ \hline
\end{tabular}
\vspace{-1em}
\end{table}

\subsection{Qualitative Analysis}
Fig.~\ref{fig_3} visualizes two results from the video retrieval task on MSR-VTT. 
In each example, the green boxes highlight the two most related class anchors, while the red boxes indicate the two least related class anchors.
In Fig.~\ref{fig_3}-(a), our model effectively associates the video with the class anchor \textit{``Holding a laptop”}, which closely aligns with the text-matched anchor \textit{``Putting a laptop somewhere”}. 
This demonstrates that class anchors successfully capture semantic cues within each input modality.
In Fig.~\ref{fig_3}-(b), we present another case where the video contains complex actions that are not explicitly described in the accompanying text. 
The video aligns with class anchors corresponding to these intricate actions, while the text aligns with anchors related to the general activities of \textit{``a man is talking on stage''}. 
This highlights the information imbalance between the modalities. 
Our model effectively handles this imbalance by aligning probability distributions in a generative manner rather than directly matching input features. 
By doing so, it addresses uncertainty and partial alignments, thus enhancing both inter-modal and intra-modal understanding through underlying semantic relationships.

\section{Conclusion}
In this work, we present CALM, Class-anchor-ALigned generative Modeling, to address modality discrepancies and information imbalance in multi-modal representation learning. 
By leveraging class anchors as prompts and aligning class probability distributions across modalities, CALM effectively bridges the modality gap and captures deeper semantic relationships between video and text. 
Our cross-modal probabilistic variational autoencoder models uncertainty and accommodates variations across modalities by reconstructing the intra-modal distribution from the inter-modal distribution. 
Extensive experiments on four benchmark datasets demonstrate that CALM consistently outperforms state-of-the-art methods, with significant improvements in out-of-domain evaluations, highlighting its superior generalization capabilities.

\section*{Acknowledgment}
This work was supported in part by the National Research Foundation of Korea (NRF) grant funded by the Korea government (MSIT) (RS-2023-00279019) and in part by the Institute of Information \& Communications Technology Planning \& Evaluation (IITP) grant funded by Korea Government (MSIT) (2021-0-01341, Artificial Intelligence Graduate School Program (Chung-Ang University)).

{\small
\bibliographystyle{ieee_fullname}
\bibliography{egbib}
}

\end{document}